\documentclass{article} 
\usepackage{iclr2016_conference,times}
\usepackage[draft]{fixme}
\usepackage[colorlinks=true, linkcolor=black, citecolor=black, filecolor=black,
  urlcolor=black]{hyperref} 
\usepackage{graphicx}
\usepackage{url}
\usepackage{tikz}

\title{Neural GPUs Learn Algorithms}

\author{{\L}ukasz Kaiser \& Ilya Sutskever\\
Google Brain \texttt{\{lukaszkaiser,ilyasu\}@google.com}
}

\newcommand\sigmoid{\sigma}
\newcommand\argmax{\mathrm{argmax}}
\newcommand\gru{\mathrm{GRU}}
\newcommand\cgru{\mathrm{CGRU}}
\newcommand\sfin{s_\mathrm{fin}}
\newcommand\floor[1]{\left \lfloor{#1}\right \rfloor}

\iclrfinalcopy

\newcommand\cgruhands[1]{
\draw (0.6, 2.9) -- (1.79, 2.9);
\draw (0.6, 2.7) -- (1.79, 2.9);

\draw (0.6, 2.9) -- (1.79, 2.7);
\draw (0.6, 2.7) -- (1.79, 2.7);
\draw (0.6, 2.5) -- (1.79, 2.7);

\draw (0.6, 2.7) -- (1.79, 2.5);
\draw (0.6, 2.5) -- (1.79, 2.5);
\draw (0.6, 2.3) -- (1.79, 2.5);

\node (cg) at (1.2, 1.5) {{#1}};

\draw (0.65, 0.3) -- (1.79, 0.1);
\draw (0.65, 0.1) -- (1.79, 0.1);

\draw (0.65, 0.3) -- (1.79, 0.5);
\draw (0.65, 0.5) -- (1.79, 0.5);
\draw (0.65, 0.7) -- (1.79, 0.5);

\draw (0.65, 0.1) -- (1.79, 0.3);
\draw (0.65, 0.3) -- (1.79, 0.3);
\draw (0.65, 0.5) -- (1.79, 0.3);
}

\newcommand\cgrupic[1]{
\draw[step=0.2] (0, 0) grid (0.6, 3.0);
\cgruhands{{#1}}
}

\begin{document}

\maketitle

\begin{abstract}
Learning an algorithm from examples is a fundamental problem that has been widely studied.
It has been addressed using neural networks too, in particular by Neural Turing Machines (NTMs).
These are fully differentiable computers that use backpropagation to learn their own programming.
Despite their appeal NTMs have a weakness that is caused by their sequential nature:
they are not parallel and are are hard to train due to their large depth when unfolded.

We present a neural network architecture to address this problem: 
the \emph{Neural GPU}. It is based on a type of convolutional gated
recurrent unit and, like the NTM, is computationally universal. Unlike the NTM,
the Neural GPU is highly parallel which makes it easier to train and efficient to run.

An essential property of algorithms is their ability to handle inputs of arbitrary size.
We show that the Neural GPU can be trained on short instances of an algorithmic task
and successfully generalize to long instances.
We verified it on a number of tasks including long addition and long multiplication
of numbers represented in binary.  We train the Neural GPU on numbers with up-to
$20$ bits and observe no errors whatsoever while testing it, even on much longer numbers.

To achieve these results we introduce a technique for training deep recurrent networks:
parameter sharing relaxation.  We also found a small amount of dropout 
and gradient noise to have a large positive effect on learning and generalization.
\end{abstract}

\section{Introduction}

Deep neural networks have recently proven successful at various tasks,
such as computer vision \citep{img12}, speech recognition \citep{dahl12}, and in other domains.
Recurrent neural networks based on long short-term memory (LSTM) cells \citep{hochreiter1997} have been
successfully applied to a number of natural language processing tasks.
Sequence-to-sequence recurrent neural networks with such cells can learn very complex tasks in an end-to-end manner,
such as translation \citep{sutskever14,bahdanau2014neural,cho2014learning}, parsing \citep{KVparse15},
speech recognition \citep{las16} or image caption generation \citep{vinyals2014show}.
Since so many tasks can be solved with essentially one model, a natural question arises:
is this model the best we can hope for in supervised learning?

Despite its recent success, the sequence-to-sequence
model has limitations. In its basic form, the entire input is encoded
into a single fixed-size vector, so the model cannot generalize to inputs much longer
than this fixed capacity. One way to resolve this problem is by using
an attention mechanism \citep{bahdanau2014neural}. This allows
the network to inspect arbitrary parts of the input in every decoding step, so
the basic limitation is removed. But other problems remain, and \citet{algo15}
show a number of basic algorithmic tasks on which sequence-to-sequence
LSTM networks fail to generalize.
They propose a stack-augmented recurrent network, and it works
on some problems, but is limited in other ways.

In the best case one would desire a neural network model able to learn arbitrarily complex algorithms given enough resources. Neural Turing Machines \citep{ntm14} have this theoretical property.
However, they are not computationally efficient because they use soft attention and
because they tend to be of considerable depth.   Their depth makes the training objective difficult to optimize and impossible to parallelize
because they are learning a sequential program.
Their use of soft attention requires accessing the entire memory in order to
simulate $1$ step of computation, which introduces substantial overhead.
These two factors make learning complex algorithms using Neural Turing Machines difficult.
These issues are not limited to Neural Turing Machines, they apply to other architectures too,
such as stack-RNNs \citep{algo15} or (De)Queue-RNNs \citep{dequeue15}.
One can try to alleviate these problems using hard attention and reinforcement
learning, but such non-differentiable models do not learn well
at present \citep{reinforceNTM15}.

In this work we present a neural network model, the \emph{Neural GPU}, that addresses the above issues.
It is a Turing-complete model capable of learning arbitrary algorithms
in principle, like a Neural Turing Machine. But, in contrast to
Neural Turing Machines, it is designed to be as parallel and as shallow as possible.
It is more similar to a GPU than to a Turing machine since it uses a smaller~number of parallel computational steps.
We show that the Neural GPU works in multiple experiments:
\begin{itemize}
\item A Neural GPU can learn \emph{long binary multiplication} from examples.
  It is the first neural network able to learn an algorithm whose run-time is superlinear
  in the size of its input. Trained on up-to $20$-bit numbers, we see
  \emph{no single error} on any inputs we tested, and we tested on numbers up-to $2000$ bits long.
\item The same architecture can also learn \emph{long binary addition} and a number of other algorithmic tasks,
  such as counting, copying sequences, reversing them, or duplicating them.
\end{itemize}

\subsection{Related Work}

The learning of algorithms with neural networks has seen a lot of interest
after the success of sequence-to-sequence neural networks on language
processing tasks \citep{sutskever14,bahdanau2014neural,cho2014learning}. An attempt has
even been made to learn to evaluate simple python programs with a pure sequence-to-sequence
model \citep{ZS14}, but more success was seen with more complex models.
Neural Turing Machines \citep{ntm14} were shown to learn a number of basic
sequence transformations and memory access patterns, and their reinforcement
learning variant \citep{reinforceNTM15} has reasonable performance on a number
of tasks as well. Stack, Queue and DeQueue networks \citep{dequeue15} were
also shown to learn basic sequence transformations such as bigram flipping or
sequence reversal.

The Grid LSTM \citep{gridLSTM15} is another powerful architecture that can learn to
multiply $15$-digit decimal numbers.
As we will see in the next section, the Grid-LSTM is quite similar to the Neural GPU --
the main difference is that the Neural GPU is less recurrent and is explicitly
constructed from the highly parallel convolution operator.

In image processing, \emph{convolutional LSTMs}, an architecture similar to the Neural GPU,
have recently been used for weather prediction \citep{convLSTMweather} and image compression
\citep{convLSTMcompress}. We find it encouraging as it hints that the Neural GPU might perform well in other contexts.

Most comparable to this work are the prior experiments with the
stack-augmented RNNs \citep{algo15}. These networks manage to learn and
generalize to unseen lengths on a number of algorithmic tasks. But, as we show
in Section~\ref{sec:res}, stack-augmented RNNs trained to add numbers up-to $20$-bit long generalize
only to $\sim \!\! 100$-bit numbers, never to $200$-bit ones, and never without error.
Still, their generalization is the best we were able to obtain without using the Neural GPU
and far surpasses a baseline LSTM sequence-to-sequence model with attention.

The quest for learning algorithms has been pursued
much more widely with tools other than neural networks.
It is known under names such as program synthesis, program induction, automatic
programming, or inductive synthesis, and has a long history
with many works that we do not cover here;
see, e.g.,~\cite{G10surv} and \cite{K10surv} for a more general perspective.

Since one of our results is the synthesis of an algorithm for
long binary addition, let us recall how this problem has been addressed
without neural networks. Importantly, there are two cases of this problem
with different complexity. The easier case is when the two numbers
that are to be added are aligned at input, i.e., if the first (lower-endian)
bit of the first number is presented at the same time as the first bit of
the second number, then come the second bits, and so on, as depicted below
for $x=9=8+1$ and $y=5=4+1$ written in binary with least-significant bit left.

\begin{center}
\begin{tabular}{|c||c|c|c|c|c|c|c|c|c|}
\hline
{\bf Input}           & 1 & 0 & 0 & 1  \\
($x$ and $y$ aligned) & 1 & 0 & 1 & 0  \\ \hline
{\bf Desired Output ($x+y$)}   & 0 & 1 & 1 & 1 \\ \hline
\end{tabular}
\end{center}

In this representation the triples of bits from $(x,\ y,\ x+y)$,
e.g., $(1,1,0)\ (0,0,1)\ (0,1,1)\ (1,0,1)$ as in the figure above, form a \emph{regular} language.
To learn binary addition in this representation it therefore suffices to find a regular
expression or an automaton that accepts this language, which can be done with a variant
of Anguin's algorithm \citep{angluin87}. But only few interesting functions have
regular representations, as for example long multiplication does not \citep{BlGr00}.
It is therefore desirable to learn long binary addition without alignment,
for example when $x$ and $y$ are provided one after another.
This is the representation we use in the present paper.

\begin{center}
\begin{tabular}{|c||c|c|c|c|c|c|c|c|c|}
\hline
{\bf Input ($x,y$)}         & 1 & 0 & 0 & 1 & + & 1 & 0 & 1 & 0 \\ \hline
{\bf Desired Output ($x+y$)} & 0 & 1 & 1 & 1 &   &   &   &   &   \\ \hline
\end{tabular}
\end{center}

\section{The Neural GPU} \label{sec:cgrn}

Before we introduce the Neural GPU, let us recall the architecture of
a \emph{Gated Recurrent Unit} (GRU) \citep{cho2014learning}.
A GRU is similar to an LSTM, but its input and state are the same
size, which makes it easier for us to generalize it later;  a highway network
could have also been used \citep{srivastava2015highway}, but it lacks the reset gate.
GRUs have shown performance similar to LSTMs on a number of tasks \citep{gruEval14,lstmEval15}.
A GRU takes an input vector $x$ and a current state vector $s$,
and outputs:
\[ \gru(x,s) \ \ =\ \ u \odot s + (1 - u) \odot \tanh(W x + U (r \odot s) + B) ,
      \ \ \textrm{where} \]
\[ u = \sigmoid(W'x + U's + B')\quad \mathrm{and}\quad r = \sigmoid(W''x + U''s + B''). \]

In the equations above, $W,W',W'',U,U',U''$ are matrices and $B,B',B''$ are bias vectors;
these are the parameters that will be learned. We write $Wx$ for a matrix-vector multiplication
and $r \odot s$ for elementwise vector multiplication.
The vectors $u$ and $r$ are called \emph{gates} since their elements are
in $[0,1]$ --- $u$ is the \emph{update} gate and $r$ is the \emph{reset} gate.

In recurrent neural networks a unit like GRU is applied at every step
and the result is both passed as new state and used to compute
the output. In a Neural GPU we do not process a new input in every step.
Instead, all inputs are written into the starting state $s_0$.
This state has 2-dimensional structure: it consists of $w \times h$
vectors of $m$ numbers, i.e., it is a 3-dimensional tensor of shape $[w, h, m]$.
This \emph{mental image} evolves in time in a way defined by a
\emph{convolutional gated recurrent unit}:
\[ \cgru(s) \ \ =\ \ u \odot s + (1 - u) \odot \tanh(U * (r \odot s) + B),
     \ \ \textrm{where} \]
\[ u = \sigmoid(U' * s + B')\quad \mathrm{and}\quad r = \sigmoid(U''*s + B''). \]

$U * s$ above denotes the \emph{convolution} of a \emph{kernel bank} $U$ with the mental
image $s$. A kernel bank is a 4-dimensional tensor of shape $[k_w, k_h, m, m]$, i.e.,
it contains $k_w \cdot k_h \cdot m^2$ parameters, where $k_w$ and $k_h$ are kernel
width and height. It is applied to a mental image $s$ of shape $[w, h, m]$ which results
in another mental image $U * s$ of the same shape defined by:
\[ U * s[x,y,i] \ \ =\ \ \sum_{u=\floor{-k_w/2}}^{\floor{k_w/2}}\sum_{v=\floor{-k_h/2}}^{\floor{k_h/2}}\sum_{c=1}^{m}
     s[x+u,y+v,c] \cdot U[u,v,c,i]. \]

In the equation above the index $x+u$ might sometimes be negative or larger than
the size of $s$, and in such cases we assume the value is $0$. This corresponds to
the standard convolution operator used in convolutional neural networks with zero padding
on both sides and stride~$1$. Using the standard operator has the advantage that it is
heavily optimized (see Section~\ref{sec:discuss} for Neural GPU performance).
New work on faster convolutions, e.g., \citet{fastconv}, can be directly used in a Neural GPU.

\begin{figure}
\begin{center}
\begin{tikzpicture}[yscale=1.0] \small

\draw[fill=gray] (0, 0) rectangle (0.2, 3.0);

\node(i1) at (-0.6, 2.9) {$i_1$};
\node (idots) at (-0.6, 1.6) {$\vdots$};
\foreach \y in {0.1, 2.9} {
  \draw (-0.35, \y) -- (0.1, \y);
}
\node(in) at (-0.6, 0.1) {$i_n$};

\node (s0) at (0.3, -0.5) {$s_0$};
\cgrupic{CGRU$_1$}
\begin{scope}[xshift=1.79cm]
\cgrupic{CGRU$_2$}
\end{scope}

\begin{scope}[xshift=3.59cm]
\cgrupic{}
\end{scope}
\node (s1) at (3.9, -0.5) {$s_1$};

\node (dots) at (5.7, 1.5) {$\dots$};
\begin{scope}[xshift=5.39cm]
\cgruhands{}
\end{scope}

\node (sp) at (7.5, -0.5) {$s_{n-1}$};
\begin{scope}[xshift=7.19cm]
\cgrupic{CGRU$_1$}
\begin{scope}[xshift=1.8cm]
\cgrupic{CGRU$_2$}
\end{scope}
\end{scope}

\draw[fill=gray] (10.8, 0) rectangle (11.0, 3.0);
\draw[step=0.2] (10.79, 0) grid (11.4, 3.0);
\node (sn) at (11.1, -0.5) {$s_n$};

\node(o1) at (12.0, 2.9) {$o_1$};
\node (idots) at (12.0, 1.6) {$\vdots$};
\foreach \y in {0.1, 2.9} {
  \draw (10 .9, \y) -- (11.75, \y);
}
\node(on) at (12.0, 0.1) {$o_n$};

\end{tikzpicture}
\end{center}

\caption{Neural GPU with $2$ layers and width $w=3$ unfolded in time.}
\label{fig:cgrn}
\end{figure}
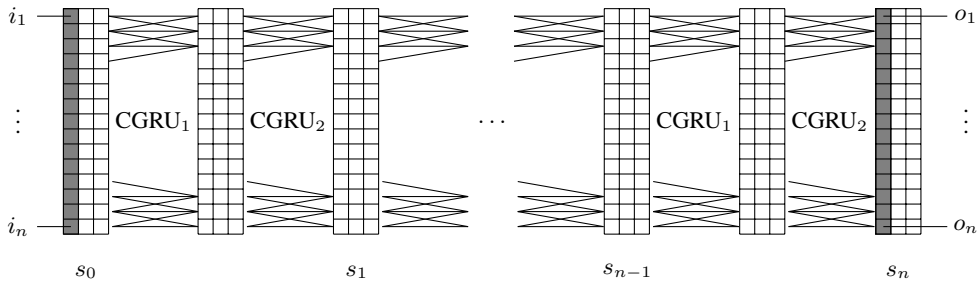

Knowing how a CGRU gate works, the definition of a $l$-layer Neural GPU is simple, as depicted in Figure~\ref{fig:cgrn}.
The given sequence $i = (i_1,\ldots,i_n)$ of $n$ discrete symbols from $\{0,\dots,I\}$ is first embedded
into the mental image $s_0$ by concatenating the vectors obtained from
an embedding lookup of the input symbols into its first column.
More precisely, we create the starting mental image $s_0$ of shape $[w,n,m]$ by using an embedding matrix $E$ of shape $[I, m]$
and setting $s_0[0,k,:] = E[i_k]$ (in python notation) for all $k=1 \dots n$  (here $i_1,\ldots,i_n$ is the input).
All other elements of $s_0$ are set to $0$. Then, we apply $l$ different CGRU gates
in turn for $n$ steps to produce the final mental image $\sfin$:
\[ s_{t+1} = \cgru_l(\cgru_{l-1} \dots \cgru_1(s_t) \dots ) \quad
     \mathrm{and} \quad \sfin = s_n. \]
The result of a Neural GPU is produced by multiplying each item in
the first column of $\sfin$ by an output matrix $O$ to obtain the logits
$l_k = O \sfin[0,k,:]$ and then selecting the maximal one: $o_k = \argmax(l_k)$. During training we use the standard loss function, i.e., we compute a softmax over the logits $l_k$ and use the negative log probability of the target as the loss.

Since all components of a Neural GPU are clearly differentiable, we can train using
any stochastic gradient descent optimizer. For the results presented in
this paper we used the Adam optimizer \citep{adam} with $\varepsilon=10^{-4}$ and gradients norm clipped to $1$.
The number of layers was set to $l=2$, the width of mental images
was constant at $w=4$, the number of maps in each mental image point was
$m=24$, and the convolution kernels width and height was always $k_w=k_h=3$.

\paragraph{Computational power of Neural GPUs.}
While the above definition is simple, it might not be immediately obvious
what kind of functions a Neural GPU can compute. Why can we expect it
to be able to perform long multiplication?
To answer such questions it is useful to draw an analogy between a Neural GPU
and a discrete 2-dimensional cellular automaton.
Except for being discrete and the lack of a gating mechanism,
such automata are quite similar to Neural GPUs.
Of course, these are large exceptions.
Dense representations have often more capacity than purely discrete
states and the gating mechanism is crucial to avoid vanishing gradients
during training. But the computational power of cellular automata is
much better understood. In particular, it is well known that a cellular
automaton can exploit its parallelism to multiply two $n$-bit numbers
in $O(n)$ steps using Atrubin's algorithm.
We recommend the online book \citep{cabook} to get an understanding
of this algorithm and the computational power of cellular automata.

\section{Experiments}

In this section, we present experiments showing that a Neural GPU
can successfully learn a number of algorithmic tasks and generalize
well beyond the lengths that it was trained on.
We start with the two tasks we focused on, long binary addition and
long binary multiplication. Then, to demonstrate the generality of the model,
we show that Neural GPUs perform well on several other tasks as well.

\subsection{Addition and Multiplication} \label{sec:res}

The two core tasks on which we study the performance of Neural GPUs are
long binary addition and long binary multiplication.
We chose them because they are fundamental tasks and because there
is no known linear-time algorithm for long multiplication.
As described in Section~\ref{sec:cgrn}, we input a sequence of
discrete symbols into the network and we read out a sequence of symbols again.
For binary addition, we use a set of $4$ symbols: $\{0,1,+,\mathrm{PAD}\}$ and for
multiplication we use $\{0,1,\cdot,\mathrm{PAD}\}$. The $\mathrm{PAD}$
symbol is only used for padding so we depict it as empty space below.

\paragraph{Long binary addition (\texttt{badd})} is the task
of adding two numbers represented lower-endian in binary
notation. We always add numbers of the same length, but we allow
them to have $0$s at start, so numbers of differing lengths can
be padded to equal size. Given two $d$-bit numbers the full
sequence length is $n=2d+1$, as seen in the example below,
representing $(1 + 4)+(2 + 4 + 8)=5+14=19=(16+2+1)$.

\begin{center}
\begin{tabular}{|c|c|c|c|c|c|c|c|c|c|}
\hline
{\bf Input} & 1 & 0 & 1 & 0 & + & 0 & 1 & 1 & 1 \\ \hline
{\bf Output} & 1 & 1 & 0 & 0 & 1 &   &   &   &   \\ \hline
\end{tabular}
\end{center}

\paragraph{Long binary multiplication (\texttt{bmul})} is the task
of multiplying two binary numbers, represented lower-endian.
Again, we always multiply numbers of the same length, but we allow
them to have $0$s at start, so numbers of differing lengths can
be padded to equal size. Given two $d$-bit numbers, the full
sequence length is again $n=2d+1$, as seen in the example below,
representing $(2 + 4) \cdot (2 + 8) = 6 \cdot 10 = 60 = 32+16+8+4$.

\begin{center}
\begin{tabular}{|c|c|c|c|c|c|c|c|c|c|}
\hline
{\bf Input} & 0 & 1 & 1 & 0 & $\cdot$ & 0 & 1 & 0 & 1 \\ \hline
{\bf Output} & 0 & 0 & 1 & 1 & 1 &   &   &   &   \\ \hline
\end{tabular}
\end{center}

\begin{table}
\centering
\small{
\begin{tabular}{|l||c||c|c|c|}
\hline
{\bf Task@Bits}  & {\bf Neural GPU} & {\bf stackRNN} & {\bf LSTM+A} \\ \hline
\texttt{badd}@20   & 100\% & 100\% & 100\% \\
\texttt{badd}@25   & 100\% & 100\% & 73\% \\
\texttt{badd}@100  & 100\% & 88\%  & 0\% \\
\texttt{badd}@200  & 100\% & 0\%   & 0\% \\
\texttt{badd}@2000 & 100\% & 0\%   & 0\% \\
\hline
\texttt{bmul}@20   & 100\% & N/A & 0\% \\
\texttt{bmul}@25   & 100\% & N/A & 0\% \\
\texttt{bmul}@200  & 100\% & N/A & 0\% \\
\texttt{bmul}@2000 & 100\% & N/A & 0\% \\
\hline
\end{tabular}
}
\caption{Neural GPU, stackRNN, and LSTM+A results on addition and multiplication.
  The table shows the fraction of test cases for which every single bit of the model's output is correct.}
\label{tab:res}
\end{table}

\paragraph{Models.}
We compare three different models on the above tasks.
In addition to the Neural GPU we include a baseline LSTM recurrent neural network with
an attention mechanism. We call this model LSTM+A as it is exactly
the same as described in \citep{KVparse15}. It is a $3$-layer
model with $64$ units in each LSTM cell in each layer, which results
in about $200$k parameters (the Neural GPU uses $m=24$ and has about $30$k paramters).
Both the Neural GPU and the LSTM+A baseline were trained using all the techniques
described below, including curriculum training and gradient noise.
Finally, on binary addition, we also include the stack-RNN model
from \citep{algo15}. This model was not trained using our training
regime, but in exactly the way as provided in its source code,
only with $nmax=41$. To match our training procedure, we ran it $729$ times
(cf. Section~\ref{sec:train}) with different random seeds and we report the best obtained result.

\paragraph{Results.}
We measure also the rate of \emph{fully correct output sequences} and
report the results in Table~\ref{tab:res}. For both tasks, we show first
the error at the maximum length seen during training, i.e., for $20$-bit
numbers. Note that LSTM+A is not able to learn long binary multiplication
at this length, it does not even fit the training data.
Then we report numbers for sizes not seen during training.

As you can see, a Neural GPU can learn a multiplication algorithm that
generalizes perfectly, at least as far as we were able to test (technical
limits of our implementation prevented us from testing much above $2000$ bits).
Even for the simpler task of binary addition, stack-RNNs work
only up-to length $100$. This is still much better than the LSTM+A baseline
which only generalizes to length $25$.

\subsection{Other Algorithmic Tasks}

In addition to the two main tasks above, we tested Neural GPUs
on the following simpler algorithmic tasks. The same 
architecture as used above was able to solve all of the tasks
described below, i.e., after being trained on sequences of length
up-to $41$ we were not able to find any error on sequences on
any length we tested (up-to $4001$).

\paragraph{Copying sequences} is the simple task of
producing on output the same sequence as on input.
It is very easy for a Neural GPU, in fact all models
converge quickly and generalize perfectly.

\paragraph{Reversing sequences} is the task of reversing
a sequence of bits, $n$ is the length of the sequence.


\paragraph{Duplicating sequences} is the task
of duplicating the input bit sequence on output twice,
as in the example below. We use the padding symbol on
input to make it match the output length. We trained
on sequences of inputs up-to $20$ bits, so outputs
were up-to $40$-bits long, and tested on inputs up-to $2000$ bits long.

\begin{center}
\begin{tabular}{|c|c|c|c|c|c|c|c|c|c|}
\hline
{\bf Input} & 0 & 0 & 1 & 1 &   &   &   &   \\ \hline
{\bf Output} & 0 & 0 & 1 & 1 & 0 & 0 & 1 & 1 \\ \hline
\end{tabular}
\end{center}

\paragraph{Counting by sorting bits} is the task
of sorting the input bit sequence on output. 
Since there are only $2$ symbols to sort, this
is a counting tasks -- the network must count how many
$0$s are in the input and produce the output accordingly,
as in the example below.

\begin{center}
\begin{tabular}{|c|c|c|c|c|c|c|c|c|c|}
\hline
{\bf Input} & 1 & 0 & 1 & 1 & 0 & 0 & 1 & 0  \\ \hline
{\bf Output} & 0 & 0 & 0 & 0 & 1 & 1 & 1 & 1 \\ \hline
\end{tabular}
\end{center}

\subsection{Training Techniques} \label{sec:train}

Here we describe the training methods that we used to improve our results.
Note that we applied these methods to the LSTM+A baseline as well, to keep
the above comparison fair. We focus on the most important elements of
our training regime, all less relevant details can be found in the code
which is released as open-source.\footnote{The code is at
\url{https://github.com/tensorflow/models/tree/master/neural_gpu}.}

\paragraph{Grid search.}
Each result we report is obtained by running a grid search over $3^6=729$
instances. We consider $3$ settings of the learning rate, initial parameters
scale, and $4$ other hyperparameters discussed below: the relaxation pull factor,
curriculum progress threshold, gradient noise scale, and dropout.
An important effect of running this grid search is also that we train
$729$ models with different random seeds every time. Usually only
a few of these models generalize to $2000$-bit numbers, but a significant
fraction works well on $200$-bit numbers, as discussed below.

\paragraph{Curriculum learning.}
We use a curriculum learning approach inspired by \citet{ZS14}.
This means that we train, e.g., on $7$-digit numbers only after crossing
a curriculum progress threshold (e.g., over $90\%$ fully correct outputs)
on $6$-digit numbers. However, with $20\%$ probability we pick a minibatch
of $d$-digit numbers with $d$ chosen uniformly at random between $1$ and $20$.

\paragraph{Gradients noise.}
To improve training speed and stability we add noise to gradients
in each training step. Inspired by the schedule from \citet{gradNoise11},
we add to gradients a noise drawn from the normal distribution
with mean $0$ and variance inversely proportional to the square root of step-number
(i.e., with standard deviation proportional to the $4$-th root
of step-number). We multiply this noise by the gradient noise scale
and, to avoid noise in converged models, we also multiply it by
the fraction of non-fully-correct outputs (which is $0$ for a perfect model).

\paragraph{Gate cutoff.}
In Section~\ref{sec:cgrn} we defined the gates in a CGRU using the sigmoid function,
e.g., we wrote $u = \sigmoid(U' * s + B')$. Usually the standard
sigmoid function is used, $\sigmoid(x) = \frac{1}{1+e^{-x}}$. We found
that adding a hard threshold on the top and bottom helps slightly in
our setting, so we use $1.2 \sigmoid(x) - 0.1$ cut to the interval $[0,1]$,
i.e., $\sigmoid'(x) = \max(0, \min(1, 1.2 \sigmoid(x) - 0.1))$.

\subsubsection{Dropout on recurrent connections}

Dropout is a widely applied technique for regularizing neural networks.
But when applying it to recurrent networks, it has been counter-productive
to apply it on recurrent connections -- it only worked when applied to
the non-recurrent ones, as reported by \citet{pham2014dropout}.

Since a Neural GPU does not have non-recurrent connections it might seem
that dropout will not be useful for this architecture.
Surprisingly, we found the contrary -- it is useful and improves
generalization. The key to using dropout effectively in this
setting is to set a small dropout rate.

When we run a grid search for dropout rates we vary them between $6\%, 9\%,$
and $13.5\%$, meaning that over $85\%$ of the values are always preserved.
It turns out that even this small dropout has large effect since we
apply it to the whole mental image $s_i$ in each step $i$.
Presumably the network now learns to include some redundancy
in its internal representation and generalization benefits from it.

Without dropout we usually see only a few models from a $729$ grid search generalize reasonably,
while with dropout it is a much larger fraction and they generalize to higher lengths.
In particular, dropout was necessary to train models for multiplication that
generalize to $2000$ bits.

\subsubsection{Parameter sharing relaxation.}

To improve optimization of our deep network we use a \emph{relaxation}
technique for shared parameters which works as follows. Instead of training
with parameters shared across time-steps we use $r$ identical sets of non-shared
parameters (we often use $r=6$, larger numbers work better but use more memory).
At time-step $t$ of the Neural GPU we use the $i$-th set if $t\ \mathrm{mod}\ r = i$.

The procedure described above relaxes the network, as it can now perform
different operations in different time-steps. Training becomes easier,
but we now have $r$ parameters instead of the single shared set we want.
To unify them we add a term to the cost function representing the distance
of each parameter from the average of this parameter in all the $r$ sets.
This term in the final cost function is multiplied by a scalar which we call
the \emph{relaxation pull}. If the relaxation pull is $0$, the network behaves
as if the $r$ parameter sets were separate, but when it is large, the cost
forces the network to unify the parameters across different set.

During training, we gradually increase the relaxation pull.
We start with a small value and every time the curriculum makes progress,
e.g., when the model performs well on $6$-digit numbers,
we multiply the relaxation pull by a relaxation pull factor.
When the curriculum reaches the maximal length we average the parameters
from all sets and continue to train with a single shared parameter~set.

This method is  crucial for learning multiplication. Without it,
a Neural GPU with $m=24$ has trouble to even fit the training set,
and the few models that manage to do it do not generalize.
With relaxation almost all models in our $729$ runs manage to fit the training data.

\section{Discussion} \label{sec:discuss}

We prepared a video of the Neural GPU trained to solve the tasks
mentioned above.\footnote{The video is available at \url{https://www.youtube.com/watch?v=LzC8NkTZAF4}}.
It shows the state in each step with values of $-1$ drawn in white,
$1$ in black, and other in gray. This gives an intuition how
the Neural GPU solves the discussed problems, e.g., it is quite
clear that for the duplication task the Neural GPU learned to move a part
of the embedding downwards in each step.

\textbf{What did not work well?}
%
For one, using decimal inputs degrades performance.
All tasks above can easily be formulated with decimal
inputs instead of binary ones. One could hope that a Neural GPU
will work well in this case too, maybe with a larger $m$.
We experimented with this formulation and our results were
worse than when the representation was binary: we did not
manage to learn long decimal multiplication. Increasing
$m$ to $128$ allows to learn all other tasks in the decimal setting.

Another problem is that often only a few models in a $729$ grid search generalize
to very long unseen instances. Among those $729$ models, there
usually are many models that generalize
to $40$ or even $200$ bits, but only a few working without error for $2000$-bit numbers.
Using dropout and gradient noise improves the reliability of training and generalization,
but maybe another technique could help even more. How could we make
more models achieve good generalization? One idea that looks natural is
to try to reduce the number of parameters by decreasing $m$.
Surprisingly, this does not seem to have any influence.
In addition to the $m=24$ results presented above we ran experiments
with $m=32,64,128$ and the results were similar.
In fact using $m=128$ we got the most models to generalize.
Additionally, we observed that ensembling a few models, just by
averaging their outputs, helps to generalize: ensembles of $5$
models almost always generalize perfectly on binary tasks.

\textbf{Why use width?}
The Neural GPU is defined using two-dimensional convolutions
and in our experiments one of the dimensions is always set to $4$.
Doing so is not necessary since a one-dimensional Neural GPU that uses
four times larger $m$ can represent every function representable
by the original one.  In fact we trained a model for long binary
multiplication that generalized to $2000$-bit numbers using
a Neural GPU with width $1$ and $m=64$.  However, the width of
the Neural GPU increases the amount of information carried in
its hidden state without increasing the number of its parameters.
Thus it can be thought of as a factorization
and might be useful for other tasks.

\textbf{Speed and data efficiency.}
Neural GPUs use the standard, heavily optimized convolution operation
and are fast. We experimented with a $2$-layer Neural GPU for $n=32$ and $m=64$.
After unfolding in time it has $128$ layers of CGRUs, each operating on $32$
mental images, each $4 \times 64 \times 64$ . The joint forward-backward step
time for this network was about $0.6$s on an NVIDIA GTX 970 GPU.

We were also surprised by how data-efficient a Neural GPU can be.
The experiments presented above were all performed using $10$k
random training data examples for each training length. Since we train
on up-to $20$-bit numbers this adds to about $200$k training examples.
We tried to train using only $100$ examples per length, so about $2000$ total
training instances. We were surprised to see that it actually worked well for
binary addition: there were models that generalized well to $200$-bit numbers
and to all lengths below despite such small training set. But we never managed
to train a good model for binary multiplication with that little training data.

\section{Conclusions and Future Work}

The results presented in Table~\ref{tab:res} show clearly that there
is a qualitative difference between what can be achieved with
a Neural GPU and what was possible with previous architectures.
In particular, for the first time, we show a neural
network that learns a non-trivial superlinear-time algorithm in
a way that generalized to much higher lengths without errors.

This opens the way to use neural networks in domains that were
previously only addressed by discrete methods, such as program synthesis.
With the surprising data efficiency of Neural GPUs it could even be possible
to replicate previous program synthesis results, e.g., \citet{K12},
but in a more scalable way. It is also interesting that a Neural GPU
can learn symbolic algorithms without using any discrete state at all,
and adding dropout and noise only improves its performance.

Another promising future work is to apply Neural GPUs to language processing tasks.
Good results have already been obtained on translation with a convolutional architecture
over words \citep{KalchbrennerB13} and adding gating and recursion, like in a Neural GPU,
should allow to train much deeper models without overfitting.
Finally, the parameter sharing relaxation technique can be applied
to any deep recurrent network and has the potential to improve RNN training in general.


\small
\begin{thebibliography}{31}
\providecommand{\natexlab}[1]{#1}
\providecommand{\url}[1]{\texttt{#1}}
\expandafter\ifx\csname urlstyle\endcsname\relax
  \providecommand{\doi}[1]{doi: #1}\else
  \providecommand{\doi}{doi: \begingroup \urlstyle{rm}\Url}\fi

\bibitem[Angluin(1987)]{angluin87}
Angluin, Dana.
\newblock Learning regaular sets from queries and counterexamples.
\newblock \emph{Information and Computation}, 75:\penalty0 87--106, 1987.

\bibitem[Bahdanau et~al.(2014)Bahdanau, Cho, and Bengio]{bahdanau2014neural}
Bahdanau, Dzmitry, Cho, Kyunghyun, and Bengio, Yoshua.
\newblock Neural machine translation by jointly learning to align and
  translate.
\newblock \emph{CoRR}, abs/1409.0473, 2014.
\newblock URL \url{http://arxiv.org/abs/1409.0473}.

\bibitem[Blumensath \& Gr{\"a}del(2000)Blumensath and Gr{\"a}del]{BlGr00}
Blumensath, Achim and Gr{\"a}del, Erich.
\newblock {Automatic Structures}.
\newblock In \emph{{Proceedings of LICS 2000}}, pp.\  51--62, 2000.
\newblock URL \url{http://www.logic.rwth-aachen.de/pub/graedel/BlGr-lics00.ps}.

\bibitem[Chan et~al.(2016)Chan, Jaitly, Le, and Vinyals]{las16}
Chan, William, Jaitly, Navdeep, Le, Quoc~V., and Vinyals, Oriol.
\newblock Listen, attend and spell.
\newblock In \emph{International Conference on Acoustics, Speech and Signal
  Processing, {ICASSP}'16}, 2016.

\bibitem[Cho et~al.(2014)Cho, van Merrienboer, Gulcehre, Bougares, Schwenk, and
  Bengio]{cho2014learning}
Cho, Kyunghyun, van Merrienboer, Bart, Gulcehre, Caglar, Bougares, Fethi,
  Schwenk, Holger, and Bengio, Yoshua.
\newblock Learning phrase representations using rnn encoder-decoder for
  statistical machine translation.
\newblock \emph{CoRR}, abs/1406.1078, 2014.
\newblock URL \url{http://arxiv.org/abs/1406.1078}.

\bibitem[Chung et~al.(2014)Chung, G{\"{u}}l{\c{c}}ehre, Cho, and
  Bengio]{gruEval14}
Chung, Junyoung, G{\"{u}}l{\c{c}}ehre, {\c{C}}aglar, Cho, Kyunghyun, and
  Bengio, Yoshua.
\newblock Empirical evaluation of gated recurrent neural networks on sequence
  modeling.
\newblock \emph{CoRR}, abs/1412.3555, 2014.
\newblock URL \url{http://arxiv.org/abs/1412.3555}.

\bibitem[Dahl et~al.(2012)Dahl, Yu, Deng, and Acero]{dahl12}
Dahl, George~E., Yu, Dong, Deng, Li, and Acero, Alex.
\newblock Context-dependent pre-trained deep neural networks for
  large-vocabulary speech recognition.
\newblock \emph{{IEEE} Transactions on Audio, Speech {\&} Language Processing},
  20\penalty0 (1):\penalty0 30--42, 2012.

\bibitem[Graves et~al.(2014)Graves, Wayne, and Danihelka]{ntm14}
Graves, Alex, Wayne, Greg, and Danihelka, Ivo.
\newblock Neural turing machines.
\newblock \emph{CoRR}, abs/1410.5401, 2014.
\newblock URL \url{http://arxiv.org/abs/1410.5401}.

\bibitem[Grefenstette et~al.(2015)Grefenstette, Hermann, Suleyman, and
  Blunsom]{dequeue15}
Grefenstette, Edward, Hermann, Karl~Moritz, Suleyman, Mustafa, and Blunsom,
  Phil.
\newblock Learning to transduce with unbounded memory.
\newblock \emph{CoRR}, abs/1506.02516, 2015.
\newblock URL \url{http://arxiv.org/abs/1506.02516}.

\bibitem[Greff et~al.(2015)Greff, Srivastava, Koutn{\'{\i}}k, Steunebrink, and
  Schmidhuber]{lstmEval15}
Greff, Klaus, Srivastava, Rupesh~Kumar, Koutn{\'{\i}}k, Jan, Steunebrink,
  Bas~R., and Schmidhuber, J{\"{u}}rgen.
\newblock {LSTM:} {A} search space odyssey.
\newblock \emph{CoRR}, abs/1503.04069, 2015.
\newblock URL \url{http://arxiv.org/abs/1503.04069}.

\bibitem[Gulwani(2010)]{G10surv}
Gulwani, Sumit.
\newblock Dimensions in program synthesis.
\newblock In \emph{Proceedings of PPDP 2010}, PPDP '10, pp.\  13--24, 2010.

\bibitem[Hochreiter \& Schmidhuber(1997)Hochreiter and
  Schmidhuber]{hochreiter1997}
Hochreiter, Sepp and Schmidhuber, J{\"u}rgen.
\newblock Long short-term memory.
\newblock \emph{Neural computation}, 9\penalty0 (8):\penalty0 1735--1780, 1997.

\bibitem[Joulin \& Mikolov(2015)Joulin and Mikolov]{algo15}
Joulin, Armand and Mikolov, Tomas.
\newblock Inferring algorithmic patterns with stack-augmented recurrent nets.
\newblock \emph{CoRR}, abs/1503.01007, 2015.
\newblock URL \url{http://arxiv.org/abs/1503.01007}.

\bibitem[Kaiser(2012)]{K12}
Kaiser, {\L}ukasz.
\newblock Learning games from videos guided by descriptive complexity.
\newblock In \emph{Proceedings of the AAAI-12}, pp.\  963--970. AAAI Press,
  2012.
\newblock URL \url{http://goo.gl/mRbfV5}.

\bibitem[Kalchbrenner \& Blunsom(2013)Kalchbrenner and
  Blunsom]{KalchbrennerB13}
Kalchbrenner, Nal and Blunsom, Phil.
\newblock Recurrent continuous translation models.
\newblock In \emph{Proceedings {EMNLP} 2013}, pp.\  1700--1709, 2013.
\newblock URL \url{http://nal.co/papers/KalchbrennerBlunsom_EMNLP13}.

\bibitem[Kalchbrenner et~al.(2016)Kalchbrenner, Danihelka, and
  Graves]{gridLSTM15}
Kalchbrenner, Nal, Danihelka, Ivo, and Graves, Alex.
\newblock Grid long short-term memory.
\newblock In \emph{International Conference on Learning Representations}, 2016.
\newblock URL \url{http://arxiv.org/abs/1507.01526}.

\bibitem[Kingma \& Ba(2014)Kingma and Ba]{adam}
Kingma, Diederik~P. and Ba, Jimmy.
\newblock Adam: {A} method for stochastic optimization.
\newblock \emph{CoRR}, abs/1412.6980, 2014.
\newblock URL \url{http://arxiv.org/abs/1412.6980}.

\bibitem[Kitzelmann(2010)]{K10surv}
Kitzelmann, Emanuel.
\newblock Inductive programming: A survey of program synthesis techniques.
\newblock In \emph{Approaches and Applications of Inductive Programming, AAIP
  2009}, volume 5812 of \emph{LNCS}, pp.\  50--73, 2010.

\bibitem[Krizhevsky et~al.(2012)Krizhevsky, Sutskever, and Hinton]{img12}
Krizhevsky, Alex, Sutskever, Ilya, and Hinton, Geoffrey.
\newblock Imagenet classification with deep convolutional neural network.
\newblock In \emph{Advances in Neural Information Processing Systems}, 2012.

\bibitem[Lavin \& Gray(2015)Lavin and Gray]{fastconv}
Lavin, Andrew and Gray, Scott.
\newblock Fast algorithms for convolutional neural networks.
\newblock \emph{CoRR}, abs/1509.09308, 2015.
\newblock URL \url{http://arxiv.org/abs/1509.09308}.

\bibitem[Pham et~al.(2014)Pham, Bluche, Kermorvant, and
  Louradour]{pham2014dropout}
Pham, Vu, Bluche, Th{\'e}odore, Kermorvant, Christopher, and Louradour,
  J{\'e}r{\^o}me.
\newblock Dropout improves recurrent neural networks for handwriting
  recognition.
\newblock In \emph{International Conference on Frontiers in Handwriting
  Recognition (ICFHR)}, pp.\  285--290. IEEE, 2014.
\newblock URL \url{http://arxiv.org/pdf/1312.4569.pdf}.

\bibitem[Shi et~al.(2015)Shi, Chen, Wang, Yeung, kin Wong, and chun
  Woo]{convLSTMweather}
Shi, Xingjian, Chen, Zhourong, Wang, Hao, Yeung, Dit-Yan, kin Wong, Wai, and
  chun Woo, Wang.
\newblock Convolutional {LSTM} network: A machine learning approach for
  precipitation nowcasting.
\newblock In \emph{Advances in Neural Information Processing Systems}, 2015.
\newblock URL \url{http://arxiv.org/abs/1506.04214}.

\bibitem[Srivastava et~al.(2015)Srivastava, Greff, and
  Schmidhuber]{srivastava2015highway}
Srivastava, Rupesh~Kumar, Greff, Klaus, and Schmidhuber, J{\"u}rgen.
\newblock Highway networks.
\newblock \emph{CoRR}, abs/1505.00387, 2015.
\newblock URL \url{http://arxiv.org/abs/1505.00387}.

\bibitem[Sutskever et~al.(2014)Sutskever, Vinyals, and Le]{sutskever14}
Sutskever, Ilya, Vinyals, Oriol, and Le, Quoc~VV.
\newblock Sequence to sequence learning with neural networks.
\newblock In \emph{Advances in Neural Information Processing Systems}, pp.\
  3104--3112, 2014.
\newblock URL \url{http://arxiv.org/abs/1409.3215}.

\bibitem[Toderici et~al.(2016)Toderici, O'Malley, Hwang, Vincent, Minnen,
  Baluja, Covell, and Sukthankar]{convLSTMcompress}
Toderici, George, O'Malley, Sean~M., Hwang, Sung~Jin, Vincent, Damien, Minnen,
  David, Baluja, Shumeet, Covell, Michele, and Sukthankar, Rahul.
\newblock Variable rate image compression with recurrent neural networks.
\newblock In \emph{International Conference on Learning Representations}, 2016.
\newblock URL \url{http://arxiv.org/abs/1511.06085}.

\bibitem[{Vinyals {\&} Kaiser} et~al.(2015){Vinyals {\&} Kaiser}, Koo, Petrov,
  Sutskever, and Hinton]{KVparse15}
{Vinyals {\&} Kaiser}, Koo, Petrov, Sutskever, and Hinton.
\newblock Grammar as a foreign language.
\newblock In \emph{Advances in Neural Information Processing Systems}, 2015.
\newblock URL \url{http://arxiv.org/abs/1412.7449}.

\bibitem[Vinyals et~al.(2014)Vinyals, Toshev, Bengio, and
  Erhan]{vinyals2014show}
Vinyals, Oriol, Toshev, Alexander, Bengio, Samy, and Erhan, Dumitru.
\newblock Show and tell: A neural image caption generator.
\newblock \emph{CoRR}, abs/1411.4555, 2014.
\newblock URL \url{http://arxiv.org/abs/1411.4555}.

\bibitem[Vivien(2003)]{cabook}
Vivien, Helene.
\newblock \emph{An Introduction to cellular automata}.
\newblock 2003.
\newblock URL
  \url{http://www.liafa.univ-paris-diderot.fr/~yunes/ca/archives/bookvivien.pdf}.

\bibitem[Welling \& Teh(2011)Welling and Teh]{gradNoise11}
Welling, Max and Teh, Yee~Whye.
\newblock Bayesian learning via stochastic gradient {L}angevin dynamics.
\newblock In \emph{Proceedings of {ICML} 2011}, pp.\  681--688, 2011.

\bibitem[Zaremba \& Sutskever(2015{\natexlab{a}})Zaremba and Sutskever]{ZS14}
Zaremba, Wojciech and Sutskever, Ilya.
\newblock Learning to execute.
\newblock \emph{CoRR}, abs/1410.4615, 2015{\natexlab{a}}.
\newblock URL \url{http://arxiv.org/abs/1410.4615}.

\bibitem[Zaremba \& Sutskever(2015{\natexlab{b}})Zaremba and
  Sutskever]{reinforceNTM15}
Zaremba, Wojciech and Sutskever, Ilya.
\newblock Reinforcement learning neural turing machines.
\newblock \emph{CoRR}, abs/1505.00521, 2015{\natexlab{b}}.
\newblock URL \url{http://arxiv.org/abs/1505.00521}.

\end{thebibliography}
\end{document}